\documentclass[10pt,twocolumn,letterpaper]{article}

\usepackage{cvpr}              % To produce the CAMERA-READY version
\usepackage{graphicx}
\usepackage{amsmath}
\usepackage{amssymb}
\usepackage{booktabs}
\usepackage{float}
\usepackage{multirow}
\usepackage{siunitx}
\usepackage{algorithm}
\usepackage{algpseudocode}
\usepackage[capitalize]{cleveref}
\crefname{section}{Sec.}{Secs.}
\Crefname{section}{Section}{Sections}
\Crefname{table}{Table}{Tables}
\crefname{table}{Tab.}{Tabs.}

\begin{document}
\title{PANeRF: Pseudo-view Augmentation for Improved Neural Radiance Fields Based on Few-shot Inputs}

\author{
Young Chun Ahn \qquad Seokhwan Jang \qquad Sungheon Park \qquad Ji-Yeon Kim \qquad Nahyup Kang\\
\\
Samsung Advanced Institute of Technology (SAIT)\\
% {\tt\small
%     \{ychun.ahn, swan.jang, sh2019.park, jiyeon31.kim, nahyup.kang\}}@samsung.com
}

\makeatletter
\g@addto@macro\@maketitle{
  \begin{figure}[H]
  \centering
  \includegraphics[width=1\textwidth]{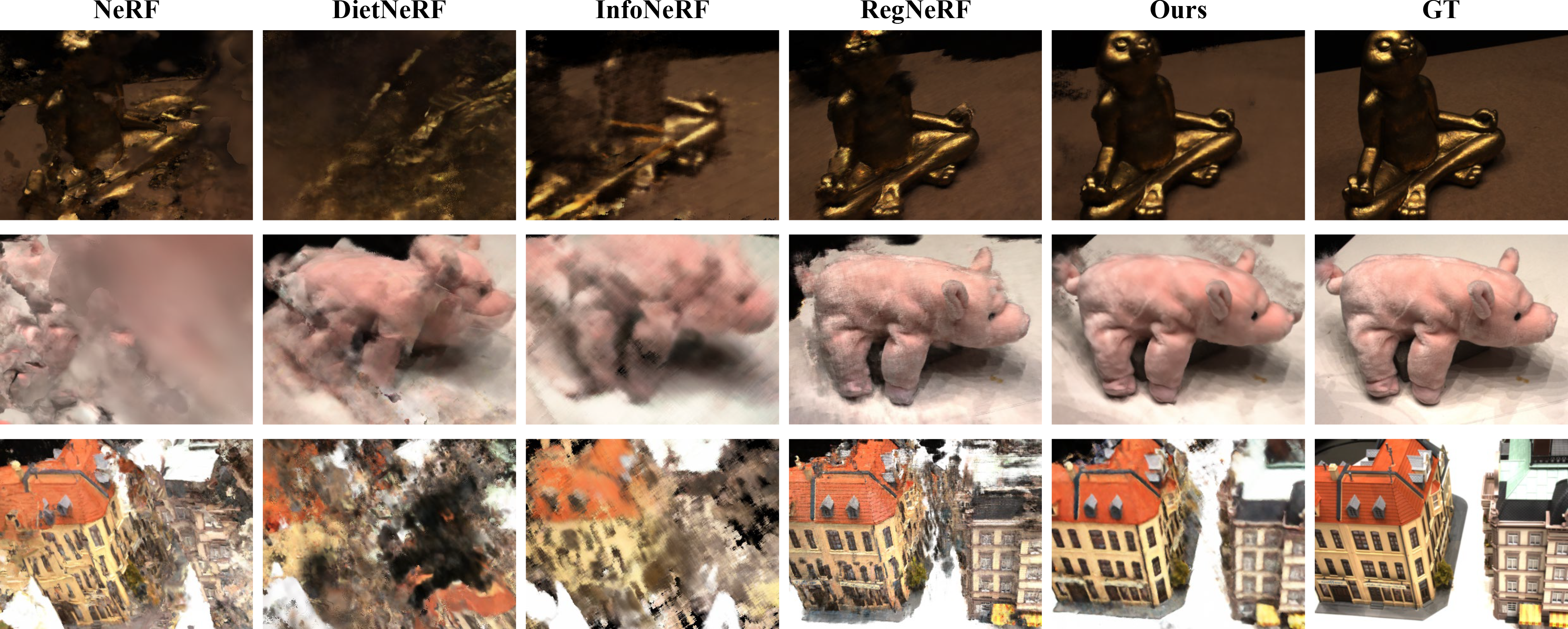}
  \setlength{\linewidth}{\textwidth}
  \setlength{\hsize}{\textwidth}
  \caption{
  View synthesis based on the DTU with a 3-view setting. We demonstrate a qualitative comparison of pseudo-view augmentation neural radiance fields (PANeRF) with other few-shot methods of \textit{Scan21}, \textit{Scan103}, and \textit{Scan110} scenes based on the DTU dataset. Although other methods experience inaccurate geometry and appearance, our approach yields high-quality rendering results with minimal artifacts.}
  \label{fig:main}
  \end{figure}
}
\maketitle
%%%%%%%%% ABSTRACT
\begin{abstract}
   \textit{The method of neural radiance fields (NeRF) has been developed in recent years, and this technology has promising applications for synthesizing novel views of complex scenes. However, NeRF requires dense input views, typically numbering in the hundreds, for generating high-quality images. With a decrease in the number of input views, the rendering quality of NeRF for unseen viewpoints tends to degenerate drastically. To overcome this challenge, we propose pseudo-view augmentation of NeRF, a scheme that expands a sufficient amount of data by considering the geometry of few-shot inputs. We first initialized the NeRF network by leveraging the expanded pseudo-views, which efficiently minimizes uncertainty when rendering unseen views. Subsequently, we fine-tuned the network by utilizing sparse-view inputs containing precise geometry and color information. Through experiments under various settings, we verified that our model faithfully synthesizes novel-view images of superior quality and outperforms existing methods for multi-view datasets.}
\end{abstract}

\section{Introduction}
\begin{figure*}[hbt!]
  \includegraphics[width=1\textwidth]{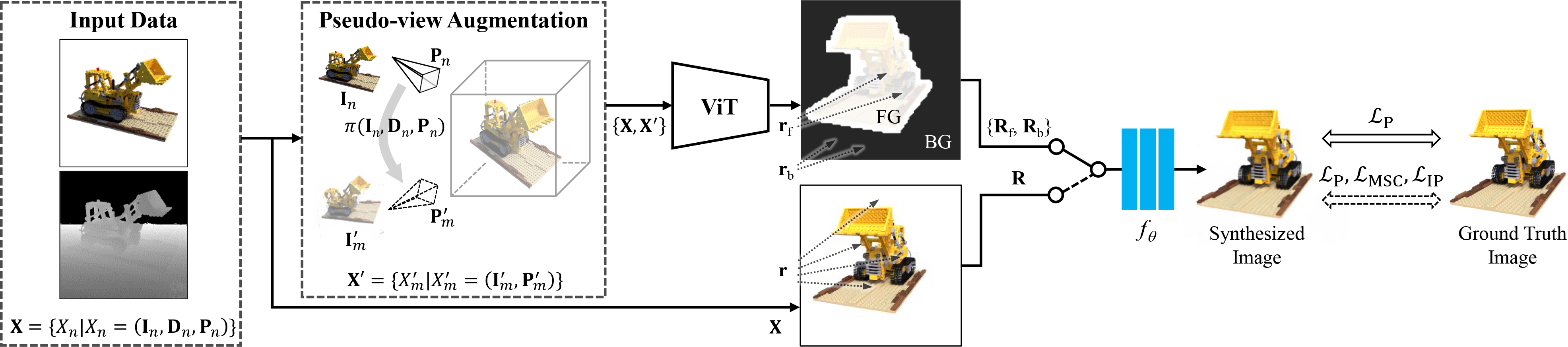}
  \caption{Overall pipeline for training PANeRF. In pseudo-view augmentation, $\pi$ denotes warping operation. $\mathbf{R}_{\text{f}}$ and $\mathbf{R}_{\text{b}}$ denote a set of 3D points on rays, $\mathbf{r}_\text{f}$ and $\mathbf{r}_\text{b}$, in the foreground and background, respectively. Note that we train our model through a two-step learning process with different losses.}
  \label{fig:overview}
\end{figure*}

Although neural radiance fields (NeRF)~\cite{NeRF} have remarkably advanced the field of neural rendering, synthesis of photo-realistic views with sparse inputs via this technology has remained a major challenge. Moreover, in real-world applications, such as AR/VR and autonomous driving, data sparsity is a major technological hurdle. When considering the use of NeRF in practical applications, achieving a reliable performance despite this limitation is essential.

To overcome the aforementioned limitation, various methods~\cite{PixelNeRF, IBRNet, DietNeRF, RegNeRF} employ a pre-trained model. In particular, these methods utilize the input-image features to construct a feature pyramid, add a ray transformer for density estimation, extract semantic information, or maximize the predicted log-likelihood. These methods yield an adequate rendering performance. However, their results feature inaccurate geometries and floating artifacts.

Another approach~\cite{InfoNeRF} proposes regularization techniques with the use of a pre-trained model.
Although this strategy enhances the performance in multi-view rendering based on adjacent sparse views such as narrow-baseline imagery, the details tend to be lost for unseen viewpoints.

In addition, certain recent works~\cite{DSNeRF, SinNeRF} have leveraged depth maps.
In one of these studies~\cite{DSNeRF}, depth information was utilized for direct supervision at the rendering view. In another study~\cite{SinNeRF}, training images were augmented through homography warping.

In accordance with this line of research, we employed a view augmentation scheme to resolve the aforementioned challenge resulting from sparse data. Similar to our approach, SinNeRF~\cite{SinNeRF} performs data augmentation using an input image and its depth map; however, the synthesized data are utilized solely via a patch-wise manner.

Specifically, although certain inaccurate regions existed owing to warping artifacts, we fully utilized the augmented images for training. To minimize the effects of warping inaccuracy, we propose computation of saliency maps through DINO-ViT~\cite{DINO}. Utilizing the saliency map, we divided training images into foreground and background regions and calculated the loss for each region separately. As a result of the advantageous features of the saliency map, all the pseudo views could be considered reliable for calculating the training loss.

Additionally, we propose a two-step learning process. First, the network was initially trained based on all the augmented images to attenuate the uncertainty corresponding to the novel views. Subsequently, the network was fine-tuned through novel regularization methods solely based on the ground-truth images. Owing to this strategy, our method yields a superior image quality despite a 1-view setting, which is considered an extreme case.

In this paper, we propose \textit{PANeRF} for synthesizing novel views based on few-shot data. The main contributions are summarized as follows:
\begin{itemize}
    \item {Our pseudo-view augmentation (PA) scheme aggregates geometric accuracy and semantic details in rendering at novel viewpoints.}
    \item {We propose novel regularization methods, namely the multi-level semantic consistency (MSC) and information potential (IP). The former method maintains semantic consistency at local and global levels, and the latter minimizes uncertainty along the rays.}
    \item {We verified the robustness of our method through numerous experiments under various conditions and achieved high-quality view synthesis superior to that rendered by existing few-shot methods.}
\end{itemize}
\section{Related Work}
{\bf \noindent Novel View Synthesis.}
For novel-view synthesis, certain explicit representations of 3D vision using voxels have been reported~\cite{Sitzmann2019DeepVoxelsLP,Lombardi2019NeuralV}, meshes~\cite{Hu2021WorldsheetWT,Shih20203DPU}, point clouds~\cite{Wiles2020SynSinEV}, or multiplane images~\cite{Tucker2020SingleViewVS,Flynn2019DeepViewVS,Mildenhall2019LocalLF,Srinivasan2019PushingTB}.
Although these methods allow for rendering of images based on the reconstructed 3D scenes, optimizing these schemes is generally a challenging task owing to their properties of discontinuity. Moreover, various studies~\cite{Niemeyer2020DifferentiableVR,Liu2019LearningTI,Yariv2020MultiviewNS} report implicit representation in which geometry and appearance can be directly learned without an explicit 3D model. In this context, neural radiance fields (NeRF)~\cite{NeRF} is a representative work that performs photo-realistic novel-view synthesis. In addition, various schemes have been introduced to improve the performance of NeRF. For example, Mip-NeRF~\cite{MipNeRF} and its extension, namely, Mip-NeRF 360\cite{MipNeRF360}, reduce aliasing by tracing a cone instead of a ray. Furthermore, Ref-NeRF\cite{RefNeRF} achieves improved rendering quality by replacing the parameterization of view-dependent radiance with the representation of reflected radiance.

{\bf \noindent Few-shot View Synthesis.}
Various methods have been suggested in the field of study to overcome limitations resulting from insufficient input data. In particular, certain approaches employ well-defined regularization techniques with or without prior models. In this perspective, PixelNeRF~\cite{PixelNeRF} utilizes a feature volume based on a convolutional neural network (CNN) encoder to train NeRF. IBRNet~\cite{IBRNet} extracts density features by accumulating image features, colors, and viewing directions of adjacent views, and transmits these features through a ray transformer to estimate a more accurate density. DietNeRF~\cite{DietNeRF} employs a vision transformer to extract semantic features for sustaining global semantic information at novel viewpoints. This scheme leverages CLIP-ViT~\cite{CLIP}, which has been trained on enormous 2D image and text data, to compensate for information scarcity. Furthermore, RegNeRF~\cite{RegNeRF} regularizes the geometry and color of patches rendered from novel views through a pre-trained normalizing-flow model.
Its geometry regularization is rendered as the patch-wise depth smoothness from unseen views. In addition, color regularization maximizes the log-likelihood of the predicted color of image patches while training by utiliziing the normalized flow model.

InfoNeRF~\cite{InfoNeRF} introduces a prior-free model that utilizes the ray entropy among seen and unseen poses. This scheme proposes a regularization technique through ray entropy minimization. The ray entropy is computed based on the density distribution by using the volume densities of the rays. Although this strategy improves the image quality of novel-view synthesis by minimizing the entropy among rays sampled in the seen and unseen views, artifacts such as blurring and cloud effects are usually rendered into the synthesized image.

Recently, a few studies have introduced the use of depth supervision for novel-view synthesis. In this regard, DSNeRF~\cite{DSNeRF} applies geometric constraint as a form of direct depth supervision; it applies the supervision using sparse 3D points obtained from structure-from-motion (SFM) such as COLMAP~\cite{COLMAP}. Moreover, depth supervision can be performed by projecting the 3D points with the corresponding camera parameters. The most recently proposed strategy in this regard, SinNeRF~\cite{SinNeRF}, addresses more extreme conditions wherein neural radiance fields are trained solely using a single view. This strategy augments the geometry-based pseudo labels near the reference view via forward image warping based on depth supervision. However, on account of the incorrect warping, the geometry and texture-guidance loss are designed in a small patch.
%% Method
\section{Method}
\Cref{fig:overview} illustrates an overview of our proposed method.
We propose a two-step learning process comprising network initialization and fine-tuning. For the network initialization, we introduce PA, a technique for expanding novel views from given input data to their surroundings via forward image warping based on depth information. In addition, we extracted their saliency maps to minimize artifacts resulting from warping inaccuracy during the training. Following the initialization, the network was fine-tuned using only the few-shot input images. For this purpose, we employed novel regularization methods. MSC was applied to maintain semantic attributes, and IP was adopted to minimize uncertainty in novel viewpoints. The relevant details are described in \cref{sec:PA}, \cref{sec:MSC}, and \cref{sec:IP}, respectively.

%% Preliminary
\subsection{Preliminary}
NeRF~\cite{NeRF} represents a 3D scene with a neural implicit function using a multi-layer perceptron (MLP). In this approach, a 3D position, $\mathbf{x} = \mathcal(x,y,z)$, and a viewing direction, $\mathbf{d}= \mathcal(\theta,\phi)$, are fed as inputs to the MLP network, and the volume density $\sigma$ and color $\mathbf{c}=\mathcal(r,g,b)$ at the 3D point are predicted accordingly.
By integrating colors and densities along the ray based on the volume rendering~\cite{volumeRendering}, the RGB color at the target pixel is rendered.
In practice, a subset of points on a ray is sampled and fed into the network to produce the rendered RGB color, as expressed in (\ref{eqn:colorNeRF}):

\begin{equation}
\hat{C}(\mathbf{r}) = \sum_{i=1}^{N}T_i(1-\exp(-\sigma_i\delta_i))\mathbf{c}_i,
\label{eqn:colorNeRF}
\end{equation}
\noindent where \textbf{r} denotes a ray, \textit{N} indicates the total number of sample points, and $\delta_i$ represents the distance between the $i^\text{th}$ and $(i+1)^\text{th}$ samples.
$T_i=\exp{(-\sum_{j=1}^{i-1}\sigma_j\delta_j)}$ symbolizes the transmittance accumulated along ray until the $i^\text{th}$ sample point. The NeRF model is optimized by minimizing the photometric loss between the ground truth and synthesized images, which can be expressed as follows:

\begin{equation}
\mathcal{L}_{\text{P}} = \sum_{\mathbf{r}\in\mathcal{R}}\left\|\hat{C}(\mathbf{r})-C(\mathbf{r})  \right\|^2,
\label{eqn:lossP}
\end{equation}
\noindent where $\mathcal{R}$ denotes a set of training rays and $C$ is a pixel color of a training image.

%% Pseudo-view Augmentation
\subsection{Pseudo-view Augmentation}
\label{sec:PA}
To address the scarcity of viewpoints, pseudo-view images are generated through homography warping, as depicted in \cref{fig:overview}.
Referring to the poses of few-shot inputs, we initially rotated the camera pose in the \textit{x}, \textit{y}, and \textit{z} axes in the range of  $\pm \alpha^\circ$, respectively.
Subsequently, warping is performed ranging from the reference camera pose $\mathbf{P}$ to the transformed pose $\mathbf{P}^\prime$. Accordingly, the transformed pixel position $\textbf{p}^{\prime}$ can be obtained through (\ref{eqn:forward_warping}) as follows: 

\begin{equation}
\mathbf{p}^{\prime} = \mathbf{K}\mathbf{T}_{\mathbf{P}\rightarrow \mathbf{P}^\prime}D(\mathbf{p})\mathbf{K}^{-1}\mathbf{p},
\label{eqn:forward_warping}
\end{equation}

\noindent where $\mathbf{K}$, $\mathbf{T}_{\mathbf{P}\rightarrow \mathbf{P}^\prime}$, and $D$ denote a camera intrinsic matrix, transformation matrix from $\mathbf{P}$ to $\mathbf{P}^\prime$, and depth map, respectively. Notably, the forward-warping scheme with scattering operation is adopted in our implementation. Specifically, a source pixel value is distributed to the target position's neighboring pixels via the scattering operation.

Moreover, the quality of the image obtained using the forward-warping scheme primarily depends on the accuracy of the depth map. To minimize this dependency, in this study, we adopted the reliable depth map estimated from the pre-trained NeRF~\cite{NeRF}. To further attenuate the warping effect, we generated saliency maps by employing DINO-ViT~\cite{DINO}, and these maps were leveraged to separately supervise pixel values of the foreground and background.

\begin{figure}[t!]
  \centering
  \includegraphics[width=0.47\textwidth]{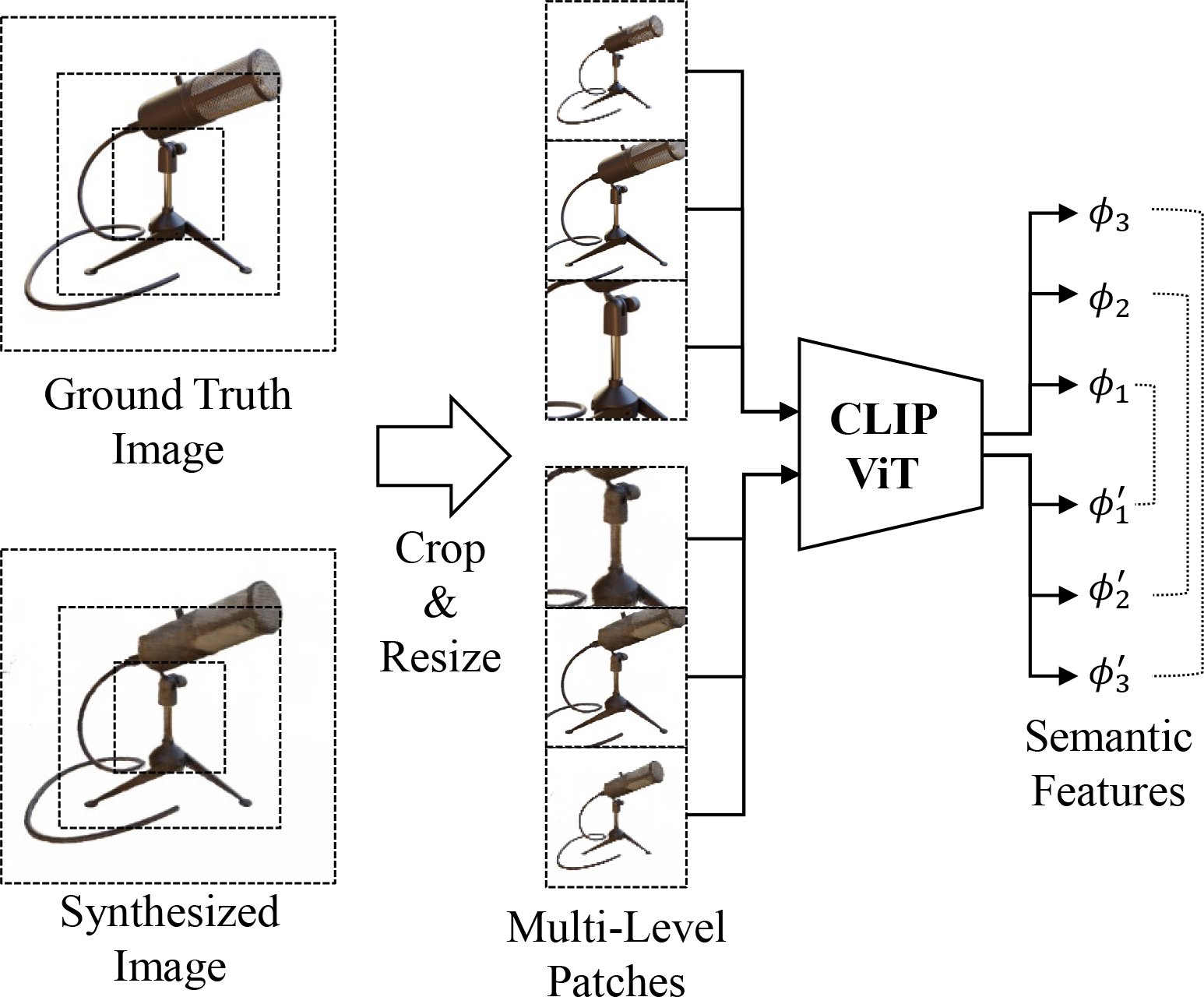}
  \caption{
  Illustration of measuring the MSC between the ground-truth and synthesized images in the case of $L=3$.
  Unlike the approach assumed in another study~\cite{DietNeRF}, we investigated the consistency at local levels as well as global levels.}
  \label{fig:MSC}
\end{figure}

\begin{figure}[t]
  \centering
  \includegraphics[width=0.47\textwidth]{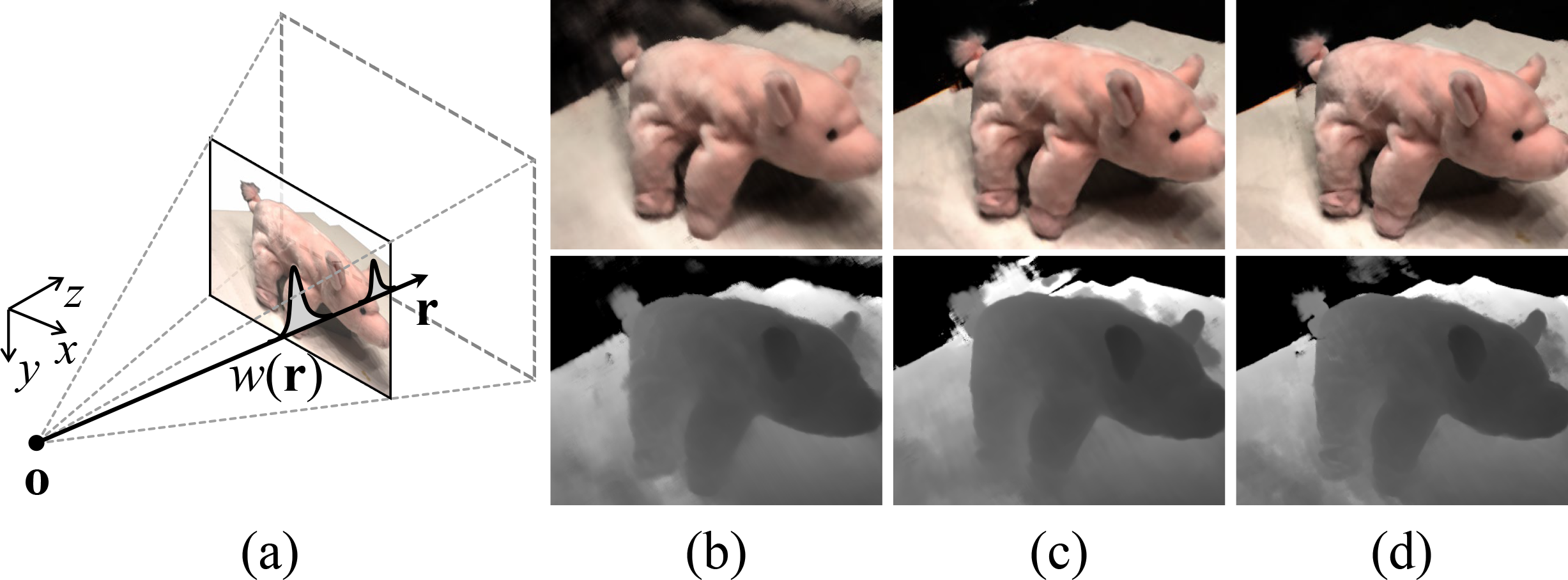}
  \caption{
 (a) Illustration of ray $\mathbf{r}$, camera position $\mathbf{o}$, and its weight distribution $w(\mathbf{r})$. The synthesized images and their depth maps obtained (b) after the initialization, (c) without IP, (d) with IP, respectively.}
  \label{fig:IP}
\end{figure}

%% Multi-level Semantic Consistency
\subsection{Multi-level Semantic Consistency}
\label{sec:MSC}
We explored semantic features from the pre-trained CLIP-ViT~\cite{CLIP} to ensure consistent semantic information for synthesizing novel views. In particular, DietNeRF~\cite{DietNeRF} extracts features using CLIP-ViT~\cite{CLIP}; however, the features are employed at a global level. In contrast, we propose the MSC that can maintain semantic attributes at both local and global levels.

To extract the embedding features at each level, we obtained patches of three different sizes by cropping an image around its center as illustrated in \cref{fig:MSC}. The patch-wise images were scaled and subsequently fed into the CLIP-ViT~\cite{CLIP}. Thereafter, the cosine similarities of the embedding features obtained from the same region were calculated, and summed up as the MSC loss. This loss is expressed as

\begin{equation}
\mathcal{L}_{\text{MSC}} = \sum_{l=1}^{L} S(\phi_l(\hat{\mathbf{I}}), \phi_l(\mathbf{I})),
\label{eqn:MSC}
\end{equation}
\noindent where $\hat{\mathbf{I}}$ and $\mathbf{I}$ denote the synthesized and the ground-truth images, respectively. Notably, $L$ indicates the number of levels, and $\phi_l$ denotes embedding features at level \textit{l}. $\mathcal{S(\cdot)}$ symbolizes a function of cosine similarity between two embedding features.

%% Information Potential
\subsection{Information Potential}
\label{sec:IP}
In addition to the MSC, we introduce a new regularization method that suppresses the uncertainty in transmittance along the rays, which results from scarce data. We observed in an experiment that a weight distribution on a ray, which is represented by multiplying the transmittance and opacity at sample point as in (\ref{eqn:colorNeRF}), tends to be concentrated on the surfaces of a scene, as illustrated in~\cref{fig:IP}(a).

Considering that the Shannon entropy~\cite{Shannon2001AMT} decreases as the probability density function sharpens, the entropy of the weight distribution will be minimized when the NeRF model is optimized on the scene. Based on this observation, we adopted IP which is derived from the Rényi quadratic entropy~\cite{xu2010renyi}. It is a convex function, which is more suitable for gradient-based optimization than the Shannon entropy, as proven for 3D-image reconstruction \cite{Jang2020HeadMC}, \cite{Kim2017CardiacIR}.
The IP with respect to the weights can be expressed by discarding the negative logarithm from the quadratic entropy as follows:

\begin{equation}
\mathcal{L}_{\text{IP}} = -\frac{1}{\left | \mathcal{R} \right |} \sum_{\mathbf{r}\in \mathcal{R}} \sum_{i=1}^N \tilde{w}_i(\mathbf{r})^{2},
\label{eqn:IP1}
\end{equation}
\noindent where 

\begin{equation}
\tilde{w}_i(\mathbf{r})=w_i(\mathbf{r}) / \sum_{j=1}^N w_j(\mathbf{r})
\label{eqn:IP2}
\end{equation}
\noindent and

\begin{equation}
w_i(\mathbf{r})=T_i(1-\exp(-\sigma_i\delta_i)).
\label{eqn:IP3}
\end{equation}
\noindent Herein, $\mathcal{R}$ denotes a set of rays. Notably, in (\ref{eqn:IP2}) and (\ref{eqn:IP3}), $w_i(\mathbf{r})$ represents a weight at the $i^{\text{th}}$ point that is sampled on ray \textbf{r}.
Moreover, IP observably increases when the weight distribution is concentrated on specific bins, and therefore, $\mathcal{L}$\textsubscript{IP} is multiplied with $-1$ to maximize its value during the training. 

As the IP is designed to complement the aforementioned tendency, the cloudy artifacts are attenuated. Furthermore, the boundary of the object is rendered clearer with the application of our regularizer IP during training, as indicated by the depth maps presented in \cref{fig:IP}.

%% Overall Objective
\subsection{Overall Objective}
The total loss for training our model is given by

\begin{equation}
\mathcal{L}_{total}=\mathcal{L}_{\text{P}} + \lambda_{1}\mathcal{L}_{\text{MSC}} + \lambda_{2}\mathcal{L}_{\text{IP}},
\end{equation}
where ${\lambda}$\textsubscript{1} and ${\lambda}$\textsubscript{2} denote parameters for balancing the data term, acting as two regularizers. \cref{alg:training} summarizes the entire training process.

\begin{figure*}[t!]
  \centering
  \includegraphics[width=1\linewidth]{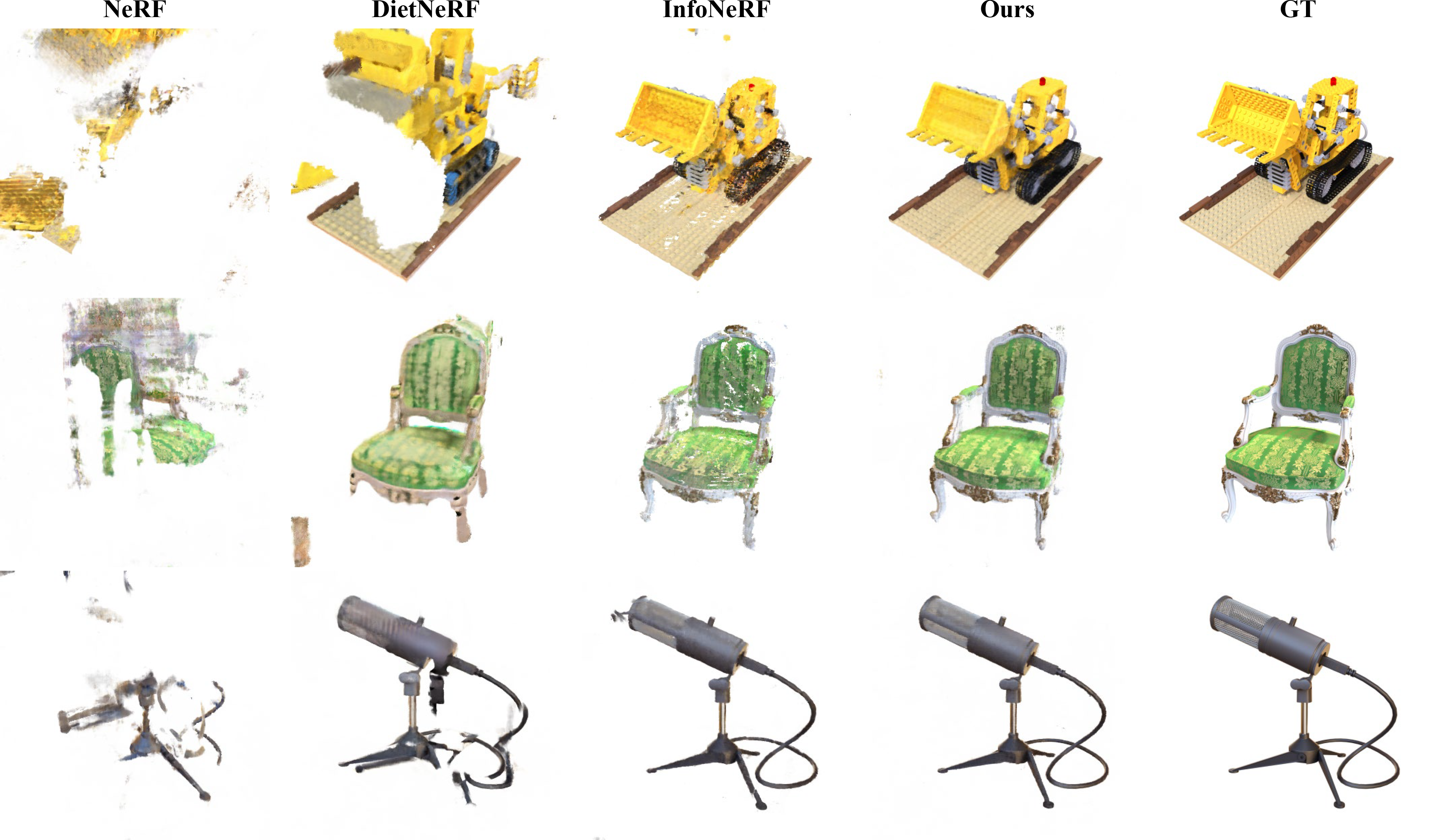}
  \caption{
  View synthesis on the Realistic Synthetic \SI{360}{\degree} with a 4-view setting. As revealed in comparison of our model to other few-shot NeRF methods, PANeRF yields a more realistic rendering.
  Our approach exhibits accurate scene geometry in \textit{Lego} and \textit{Chair}, generating superior details, as observed in \textit{Mic}.}
  \label{fig:360}
\end{figure*}

%% Algorithm
\begin{algorithm}[t!]
\caption{Training PANeRF}\label{alg:training}
\begin{algorithmic}[1]
\State \textbf{Input:} Input data $\mathbf{X}=\{X_n|X_n=(\mathbf{I}_n, \mathbf{D}_n, \mathbf{P}_n)\}$, loss weights $\lambda_1, \lambda_2$, learning rate $\eta$
\State \textbf{Output:} Radiance field $f_\theta$ with the optimum $\theta$
\State Generate $\mathbf{X}^\prime=\{X^\prime_m|X^\prime_m=(\mathbf{I}^\prime_m, \mathbf{P}^\prime_m)\}$ via \newline Pseudo-view Augmentation
\State Initialize $\theta = \theta_0$
\For {$\mathbf\xi \subset \{\mathbf{X}, \mathbf{X}^\prime$\}}  \hfill \Comment{\textit{Initialization}}
    \State Sample rays $\mathbf{r} \in \mathcal{R}$
    \State Calculate $\mathcal{L}_\text{P}$ using \cref{eqn:lossP}
    \State $\mathcal{L}_{\text{total}} \leftarrow \mathcal{L}_P$
    % \State Update $\theta \leftarrow Adam(\theta,\eta,\nabla_{\theta}\mathcal{L}_{total})$
    \State $\theta \leftarrow \theta-\eta\nabla_{\theta}\mathcal{L}_{total}$
\EndFor
\For {$\mathbf\xi \subset \mathbf{X}$} \hfill \Comment{\textit{Fine-tuning}}
    \State Sample rays $\mathbf{r} \in \mathcal{R}$
    \State Calculate losses using Eqs. (\ref{eqn:lossP}), (\ref{eqn:MSC}), and (\ref{eqn:IP1})
    % \State Calculate losses using \cref{eqn:lossP}, \cref{eqn:MSC}, and \cref{eqn:IP1}
    \State $\mathcal{L}_\text{total} \leftarrow \mathcal{L}_\text{P} + \lambda_1\mathcal{L}_\text{MSC} + \lambda_2\mathcal{L}_\text{IP}$
    \State $\theta \leftarrow \theta-\eta\nabla_{\theta}\mathcal{L}_{total}$
\EndFor
\end{algorithmic}
\end{algorithm}
\section{Experiments}
\subsection{Datasets and Evaluations}
{\bf \noindent Datasets.}
We utilized the Realistic Synthetic \SI{360}{\degree}~\cite{NeRF} dataset, which is one of the most commonly used benchmarks for novel-view synthesis. This dataset includes 8 synthetic scenes generated by natural non-Lambertian materials with complex geometries. Each scene contains 400 images, which are captured using various camera poses, covering from the upper hemisphere to the entire sphere. In addition, the DTU dataset~\cite{DTU} was employed for real-world benchmark. This dataset comprises multi-view stereo images, including those of various real-world objects, captured under calibrated camera environments. Each scene contains 49 images captured through different camera poses.

{\bf \noindent Evaluation Protocols.}
To evaluate our results on the Realistic Synthetic \SI{360}{\degree}, we followed the evaluation protocol of InfoNeRF\cite{InfoNeRF}. Our model was trained with 4 out of 100 training images, and their depth maps were used by training \cite{torchNeRF} for each scene. Subsequently, we evaluate the model with 200 testing images. In particular, 5 different sets of randomly selected 4 viewpoints were utilized to train our model, and the average performance was reported as the final result. Moreover, we evaluated our results on the DTU dataset. For a fair comparison with prior works~\cite{RegNeRF, SinNeRF}, we adopted their protocols.
Firstly, to compare with RegNeRF~\cite{RegNeRF}, we performed experiments on 15 scenes. Among 49 images of each scene, we trained our model with 3 and 6 images and evaluated the synthesized results using 25 images along with object masks. Depth maps for the PA were obtained by training \cite{torchNeRF}.
Secondly, we compared our method with SinNeRF~\cite{SinNeRF}. Experiments were performed on 19 scenes. Considering that \cite{SinNeRF} aims to train using a single view, we selected camera ID 2 as the reference view for training and tested our model with 10 images captured at views in close proximity of the reference view. For this comparison, depth maps provided by \cite{SinNeRF} were leveraged.

{\bf \noindent Evaluation Metrics.}
To evaluate our approach, the average peak signal-to-noise ratio (PSNR), structural similarity index measure (SSIM)\cite{SSIM}, and learned perceptual image patch similarity (LPIPS)~\cite{LPIPS} were measured. In addition, we report the geometric mean of MSE, $\sqrt{1-\text{SSIM}}$, and LPIPS, as conducted in a study~\cite{RegNeRF}.

\subsection{Implementation Details}
We implemented our method using the PyTorch version of the NeRF\cite{torchNeRF}.
Our model was trained for 10k iterations for the network initialization and 40k iterations for fine-tuning.
For the PA, the range of the rotation angles $\alpha$ was set to \SI{30}{\degree}, along with an interval of \SI{5}{\degree} between views. Thus, the number of images generated by referring to a source view was determined by $2,196(=\{(30\times2)/5+1)\}^3-1)$. We implemented the forward-warping scheme with the linear-scattering operation based on CUDA. In all of the experiments, the proposed model was trained and evaluated on a single NVIDIA A100 GPU.
\begin{table}
    \caption{Quantitative results on the Realistic Synthetic \SI{360}{\degree} with 4-view reconstruction. Notably we referred to Tab. 1 reported in InfoNeRF~\cite{InfoNeRF} for comparison with other models. The optimal results are marked in bold, and the secondary results are underlined.}
    \label{table:360}
    \centering
    \begin{tabular}{c c c c}
    \hline
    Model & PSNR($\,\uparrow$)\, & SSIM($\,\uparrow$)\, & LPIPS($\,\downarrow$)\, \\
    \hline\hline
    NeRF\cite{NeRF} & 15.93 & 0.780 & 0.320 \\
    PixelNeRF\cite{PixelNeRF} & 16.09 & 0.738 & 0.390 \\
    DietNeRF\cite{DietNeRF} & 16.06 & 0.793 & 0.306 \\
    InfoNeRF\cite{InfoNeRF} & \underline{18.65} & \underline{0.811} & \underline{0.230} \\
    Ours & \textbf{22.13} & \textbf{0.839} & \textbf{0.156} \\
    \hline
    \end{tabular}
\end{table}

\section{Results}
\subsection{Realistic Synthetic 360}
On the Realistic Synthetic \SI{360}{\degree} dataset, we compare our method with InfoNeRF~\cite{InfoNeRF}, whose NeRF model is trained using 4 input views. \Cref{fig:360} depicts the rendering images on the 3 different scenes, respectively. The experiments of the prior works, NeRF~\cite{NeRF}, DietNeRF~\cite{DietNeRF}, and InfoNeRF~\cite{InfoNeRF}, are reproduced by following the evaluation protocol reported in the literature \cite{InfoNeRF}. NeRF tends to degenerate when trained with sparse views. PixelNeRF and DietNeRF achieve relatively superior results via prior models, but the improvement is insignificant. InfoNeRF incorporates prior-free model and yield better results. Nevertheless, it does not yield competitive results. Our method exhibits accurate geometry and appearance representation in the synthesized results, whereas certain artifacts such as the cloud effect are clearly observed in the images rendered by the previous works. Furthermore, the quantitative evaluation indicates a significant improvement on the performance compared to that of the existing methods, as indicated in \cref{table:360}.

\begin{table*}[h!]
    \caption{Quantitative results on the DTU dataset with 3, 6, and 9 views. Notably we referred to Tab. 1 reported in \cite{RegNeRF} for the previous methods except for those of InfoNeRF~\cite{InfoNeRF} whose results are reproduced.}
    \label{table:DTU3view}
    \resizebox{\textwidth}{!}{
        \begin{tabular}{c|ccc|ccc|ccc|ccc} 
        \hline
        \multirow{2}{*}{Model} & \multicolumn{3}{c|}{PSNR($\,\uparrow)\,$}             & \multicolumn{3}{c|}{SSIM($\,\uparrow)\,$}             & \multicolumn{3}{c|}{LPIPS($\,\downarrow)\,$}            & \multicolumn{3}{c}{Average($\,\downarrow)\,$}            \\
                               & 3-view         & 6-view         & 9-view & 3-view         & 6-view         & 9-view & 3-view         & 6-view         & 9-view & 3-view         & 6-view         & 9-view  \\ 
        \hline\hline
        Mip-NeRF~\cite{MipNeRF}               & 8.68           & 16.54          & 23.58  & 0.571          & 0.741          & 0.879  & 0.353          & 0.198          & 0.092  & 0.323          & 0.148          & 0.056   \\
        DietNeRF~\cite{DietNeRF}               & 11.85          & 20.63          & 23.83  & 0.633          & 0.778          & 0.823  & 0.314          & 0.201          & 0.173  & 0.243          & 0.101          & 0.068   \\
        InfoNeRF~\cite{InfoNeRF}               & 14.13          & 19.84          & 21.78  & 0.626          & 0.713          & 0.741  & 0.314          & 0.265          & 0.246  & 0.213          & 0.128          & 0.108   \\
        RegNeRF~\cite{RegNeRF}                & \underline{18.89}          & \underline{22.20}          & 24.93  & \underline{0.745}          & \underline{0.841}          & 0.884  & 0.190          & \textbf{0.117} & 0.089  & \underline{0.112}          & \underline{0.071}          & 0.047   \\
        Ours                   & \textbf{24.16} & \textbf{26.60} & -      & \textbf{0.829} & \textbf{0.856} & -      & \textbf{0.163} & \underline{0.149}          & -      & \textbf{0.072} & \textbf{0.056} & -       \\
        \hline
        \end{tabular}
    }
\end{table*}

\begin{table}
    \caption{Quantitative results on the DTU dataset with a 1-view setting. Notably we referred to Tab. 2 reported in SinNeRF~\cite{SinNeRF} for comparison with other models.}
    \label{table:DTU1view}
    \resizebox{0.47\textwidth}{!}{
        \begin{tabular}{cccc} 
        \hline
        Model     & PSNR($\,\uparrow$)\, & SSIM($\,\uparrow$)\, & LPIPS($\,\downarrow$)\,  \\ 
        \hline\hline
        DSNeRF~\cite{DSNeRF}   & 12.17                                           & 0.41                                            & 0.6493                                              \\
        DietNeRF~\cite{DietNeRF}  & 12.84                                           & 0.44                                            & 0.6469                                              \\
        PixelNeRF~\cite{PixelNeRF} & 12.06                                           & 0.42                                            & 0.6471                                              \\
        SinNeRF~\cite{SinNeRF}   & \underline{16.52}                                           & \underline{0.56}                                            & \underline{0.5250}                                              \\
        Ours      & \textbf{16.61}                                           & \textbf{0.59}                                            & \textbf{0.4001}                                              \\
        \hline
        \end{tabular}
    }
\end{table}

\subsection{DTU}
In addition to the synthetic data, we compared our method with prior works~\cite{RegNeRF, SinNeRF} on the DTU dataset. \Cref{fig:main} illustrates the qualitative results on the 3 different scenes with the model trained through a 3-view setting. For the qualitative comparison, we reproduced the results of NeRF~\cite{NeRF}, DietNeRF~\cite{DietNeRF}, and InfoNeRF~\cite{InfoNeRF} by following the evaluation protocol reported in the study \cite{RegNeRF}. Notably, we referred to RegNeRF~\cite{RegNeRF} for their qualitative results.
Our method is superior to the other works in all aspects such as color, detail, and geometry, as shown in \cref{fig:main}.
In particular, as expressed in detail in the figure, the face of the statue and the tail of the pig doll are clearly expressed, respectively. In addition, the structure of the buildings is well restored and its color is correctly generated. This tendency is readily observed in the quantitative analysis as well, as demonstrated in \cref{table:DTU3view}.
Our method outperforms the other methods in terms of all of the metrics, considering the 3-view results.
Moreover, a significant difference exists among the evaluation metrics except for LPIPS at 6 input views. Notably, the performance of the proposed method on PSNR exceeded those of the other methods using 9 input views despite the fact that our model was trained using 6 views. Both the qualitative and quantitative results validate that our method based on the view extension functions optimally.

\begin{figure}
  \includegraphics[width=0.47\textwidth]{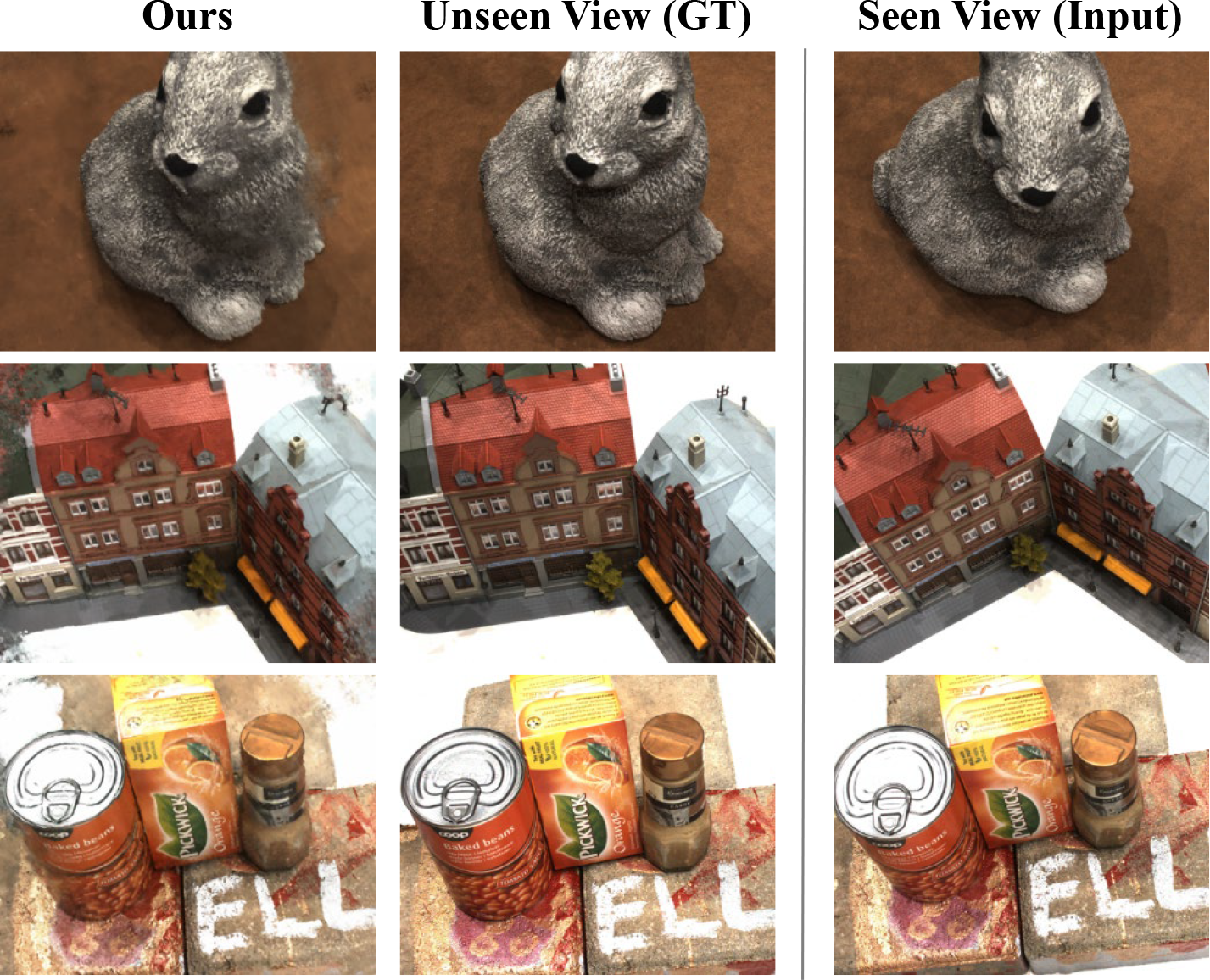}
  \caption{View synthesis on the DTU dataset with a single view.}
  \label{fig:DTU1view}
\end{figure}

Furthermore, although our model was trained using a single image, it exhibited comparable rendering quality. \Cref{fig:DTU1view} demonstrates the corresponding results.
Considering the single view (right) for training, we reveal the view-synthesis results (left) from an unseen viewpoint (middle). Our method yields the results corresponding to the correct geometry and fine details, achieving superior quantitative results. Especially significant difference in results is observed in terms of  the LPIPS, as listed in \cref{table:DTU1view}.

%%%%%%%%%%%%%%%%%%%%%%%%%%%%%%%%%%%%%%%%%%%%%%%%%%%%%%%%%%%%%

\begin{figure*}
  \includegraphics[width=1\linewidth]{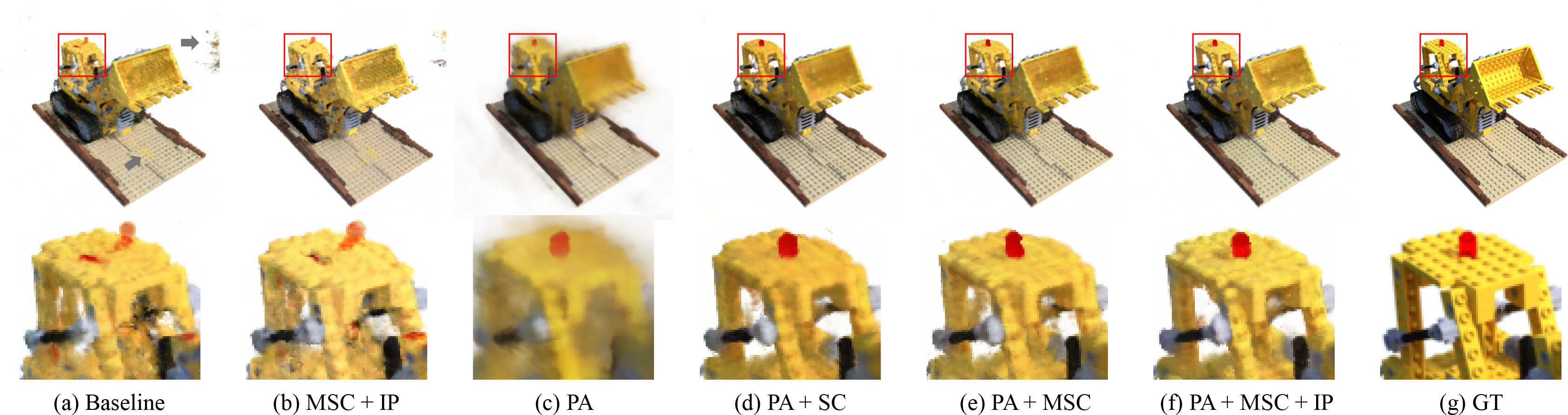}
  \caption{Results of our ablation study on the \textit{Lego} scene of the Realistic Synthetic \SI{360}{\degree}.
  The second row presents the enlarged images of the regions marked as a red box in the first row.
  (a) NeRF~\cite{NeRF} trained with a 4-view setting as our baseline.
  (b) Proposed regularization methods.
  (c) Psuedo-view augmentation only.
  (d) Add Semantic Consistency (SC) loss proposed in \cite{DietNeRF}.
  (e) MSC loss.
  (f) Our proposed method.
  (g) Ground truth.}
  \label{fig:Ablation1}
\end{figure*}

\subsection{Analysis}
{\bf \noindent Effect of the Pseudo-view Augmentation.}
As discussed in \cref{sec:PA}, our method trains the NeRF by expanding the insufficient views through PA. In \cref{fig:Ablation1}, the results suggest that PA can yield a reliable initial point in terms of optimization. By comparing the results presented in (c) with (a) and (b) in \cref{fig:Ablation1}, we discover that the PA facilitates prediction of the correct geometry and semantic information. Moreover, the proposed regularization methods along with PA can successfully minimize uncertainties; therefore, the high-quality of synthesized images can be synthesized. These results obtained via quantitative analysis validate the impact of the PA scheme, as indicated in \cref{table:Ablation}.

\begin{table}
    \caption{Ablation study for the proposed methods on the \textit{Lego} scene with a 4-view setting.}
    \label{table:Ablation}
    \resizebox{0.47\textwidth}{!}{
        \begin{tabular}{cccc|ccc}
            \multicolumn{4}{c|}{PANeRF} & \multirow{2}{*}{PSNR($\,\uparrow$)\,} & \multirow{2}{*}{SSIM($\,\uparrow$)\,} & \multirow{2}{*}{LPIPS($\,\downarrow$)\,}  \\
            PA & SC & MSC & IP          &                         &                         &                           \\ 
            \hline\hline
              &    &    &              & 20.07                   & 0.781                    & 0.190                    \\
              &    & \checkmark   & \checkmark             & 20.25                   & 0.792                    & 0.178                    \\
            \checkmark  &    &     &             & 18.75                   & 0.750                    & 0.271                    \\
            \checkmark  & \checkmark  &     &             & 22.47                   & 0.833                    & 0.149                    \\
            \checkmark  &    & \checkmark   &             & \underline{22.58}                   & \underline{0.836}                    & \underline{0.146}                    \\
            \checkmark  &    & \checkmark   & \checkmark           & \textbf{22.82}                   & \textbf{0.839}                    & \textbf{0.144}                    \\
            \hline
        \end{tabular}
    }
\end{table}

{\bf \noindent Benefit of the Regularizers.}
The proposed regularizers of MSC and IP improved the performance and contributed to the synthesis of fine details. The results in \cref{fig:Ablation1} validate that the application of the MSC generates more details than the SC. In addition, IP on top of PA and MSC, supports the proposed method in achieving the best performance.

{\bf \noindent Robustness of Our Approach.}
We performed the experiments in various environments. For the DTU dataset, our approach outperforms other methods even with 3 input views, as demonstrated in \cref{table:DTU3view}. Furthermore, even under the more challenging setting, our model is capable of achieving the high-quality view synthesis, as depicted in \cref{fig:DTU1view}.
Furthermore, we addressed the narrow-baseline setting, which is considered a difficult task. We obtained only a few viewpoints with the reported narrow view ranges in the setting. As a result, our model trained under this setting outperforms InfoNeRF in all metrics, as listed in \cref{table:LEGO_NarrowBaseline}. The synthesized results are provided in the supplementary material.

\begin{table}
    \caption{Quantitative results of the \textit{Lego} scene on the Realistic Synthetic \SI{360}{\degree} with the narrow-baseline training images.}
    \label{table:LEGO_NarrowBaseline}
    \centering
    \begin{tabular}{c c c c}
    \hline
    Model & PSNR($\,\uparrow$)\, & SSIM($\,\uparrow$)\, & LPIPS($\,\downarrow$)\, \\
    \hline\hline
    InfoNeRF~\cite{InfoNeRF} & \underline{18.41} & \underline{0.775} & \underline{0.199} \\
    Ours & \textbf{21.50} & \textbf{0.813} & \textbf{0.156} \\
    \hline
    \end{tabular}
\end{table}
\begin{figure}
  \includegraphics[width=0.47\textwidth]{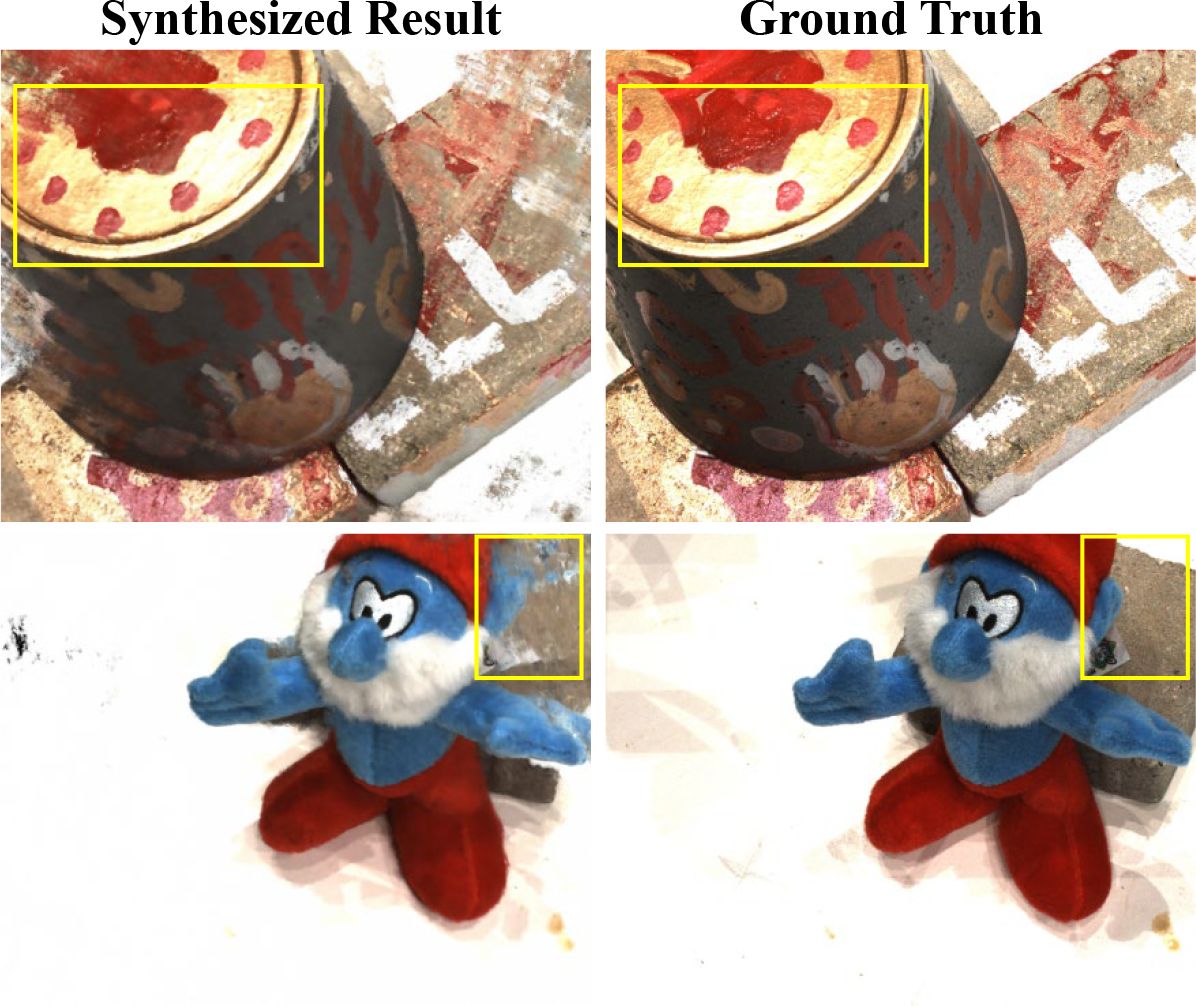}
  \caption{Limitations of our approach. The synthesized images are obtained from training our model on the DTU with 1-view.}
  \label{fig:Limitations}
\end{figure}
\section{Conclusion}
We proposed a model capable of synthesizing high-quality novel views with only sparse inputs. Through the PA scheme, we extended unseen views around the reference viewpoints for an improved learning of the geometry and appearance. Additionally, by introducing two novel regularization methods, the MSC and IP, our model could synthesize superior images featuring precise structures and fine details. The MSC maintained semantic consistency at the local and global levels, and the IP attenuated the uncertainty resulting from data sparsity. We validated the performance of the proposed method through qualitative and quantitative analysis, respectively, demonstrating that our approach outperformed the previous methods on synthetic and real-world datasets.

{\bf \noindent Limitations and Future Work.}
In the real-world dataset, the lighting effect varied according to the camera poses, as observed in the DTU dataset. As we utilized few-shot inputs, the synthesized images could not represent the exact appearance reflecting the lighting condition. Therefore, our model could not produce an accurate appearance in a view with light reflection, as illustrated in the first view in \cref{fig:Limitations}. In addition, our model was limited in generating semantic properties of the occluded regions when trained on an extreme setting, such as a single-input view. Although we could extend the views through the PA scheme, a small amount of artifact was manifested in a part of the boundary, as depicted in the second view in \cref{fig:Limitations}. We shall address these limitations in future research.

%%%%%%%%% REFERENCES
{\small
\bibliographystyle{ieee_fullname}
\bibliography{egbib}
}

\end{document}